# Repairing and Inpainting Damaged Images using Diffusion Tensor

**Faouzi Benzarti, Hamid Amiri**

**Signal, Image Processing and Pattern Recognition Laboratory**
**TSIRF (ENIT)-TUNISIA**
benzartif@yahoo.fr  Hamidlamiri@yahoo.com

**Abstract**
The Removing or repairing the imperfections of a digital images or videos is a very active and attractive field of research belonging to the image inpainting technique. This later has a wide range of applications, such as removing scratches in old photographic image, removing text and logos or creating cartoon and artistic effects. In this paper, we propose an efficient method to repair a damaged image based on a non linear diffusion tensor. The idea is to track perfectly the local geometry of the damaged image and allowing diffusion only in the isophotes curves direction. To illustrate the effective performance of our method, we present some experimental results on test and real photographic color images.

## 1. Introduction

Image inpainting technique is the process to reconstruct or filling-in missing or damage parts of image or video in a way that is undetectable to the casual observer. This technique has many applications such as: removing scratches in old photographic images and films or recovering lost blocks in the coding and transmission of images (e.g. streaming video) and also removing logos in videos. The process of inpainting can be viewed as an intelligent interpolation of adjacent pixels in the regions surrounding the areas to be recovered. Several techniques and methods have been proposed and can be divided into three categories: Structure inpainting, Texture inpainting and Completion inpainting. The structure inpainting usually involves solving PDE which is focused on the continuity of the geometrical structure of an image. In the first successful effort in this area Bertalmio and al [1] propose propagating information from the outside of the area along the level lines isophotes (lines of equal gray values). Other works introduced ideas from computational fluid dynamics into the image inpainting problem [2][22]. They showed correspondences between concepts of fluid dynamics and concepts of variational inpainting by using Navier-Stokes equations. Chan and Shen [5][6] proposed using Total Variation TV and Curvature-Driven Diffusion CDD [20]. The TV inpainting is very effective for small area which employs anisotropic diffusion based on the contrast of the isophotes. The Curvature-Driven Diffusion (CDD) model extended the TV algorithm to take into account also geometric information of isophotes when defining the strength of the diffusion process, thus allowing the inpainting to proceed over larger areas. Although some fast algorithms are developed such as Oliveaira [13] which uses convolution operator and Telea [15] proposed a fast marching algorithm. The second category of image inpainting involves texture inpainting or texture synthesis scheme which generates the target region with available sample textures from its surroundings. It is specifically useful for the images with large texture areas. Several methods were developed for synthesizing textures, including statistical and image based nonparametric sampling methods. Efros and Leung [8] adopted a nonparametric approach of "growing texture" from an initial seed. They modeled the texture as a Markov Random Field (MRF). In this model, the texture is considered to be the result of a local and stationary random MRF process. Later, exemplar-based methods were developed which resemble the non-parametric sampling method. One of the most works in this domain is the exemplar-based image inpainting of Criminisi and *al.* [7]. They exploit a patch based algorithm, in which the filling order is decided by a predefined priority function to ensure that the linear structures will propagate before texture filling to preserve the connectivity of object boundaries. The third category concerns completion inpainting which is a more general problem of filling large textured holes in images. This technique combines structure and texture simultaneously to fill image gaps with better quality and efficiency [21][3][11][4]. In this paper we focus our work on the structure inpainting by using the non linear diffusion tensor [9] which have proven their effectiveness in several areas such as: texture segmentation, motion analysis and corner detection [10]. The diffusion tensor provides a more powerful description of local pattern images better than a simple gradient. Based on its eigenvalues and the corresponding eigenvectors, the tensor allows propagating information surrounding a region of interest in a coherent direction along isophotes (level lines of equals graylevels) avoiding the smoothness across it.

This paper is organized as follows. In Section 2, we introduce the problem statement involving structure inpainting. In Section 3, we present the non linear diffusion PDEs forrmalism and its mathematical concept. Section 4 focuses on the diffusion tensor approach. The proposed method is presented in section 5. Numerical experiments and results on test and real photographic images are shown in Section 6.

## 2. Problem statement

As introduced previously, the structure inpainting usually involves solving PDE which is focused on the continuity of the geometrical structure of an image. Let $\Omega \subset D$ stands for the region to be inpainted as shown in figure 1, and $\partial\Omega$ its boundary. The objective is to fill the hole $\Omega$ with an appropriate grayvalues by interpolating the data located at the neighborhood in the surrounding region $\partial\Omega$ (e.g. boundaries).

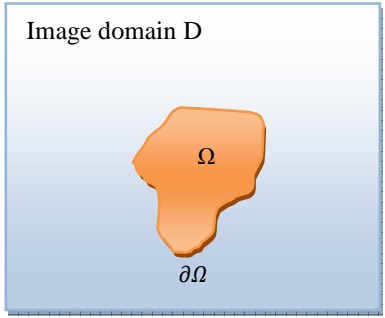

**Fig. 1** Principle of inpainting

The idea is to interpolate isophotes lines arriving at $\partial\Omega$ while maintaining the angle of arrival. The isophotes are proceeded to be drawn inward as shown in figure 2, while curving the prolongation lines progressively to prevent them from crossing one other.

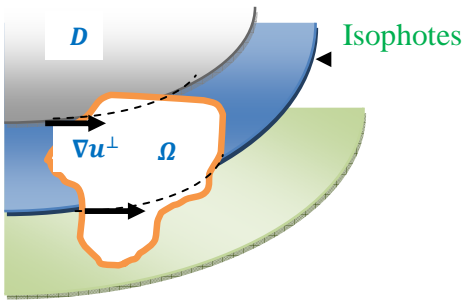

**Fig. 2** Isophotes curves prolongation

Mathematically, the direction of the isophotes can be interpreted as the orthogonal direction of the gradient $\nabla u^\perp = (-u_y, u_x)$, where $u(x, y, t)$: is the grayscaled intensity image defined in $D \in R^2 \rightarrow R$; and the smoothness can be interpreted as the Laplacian of the image $\Delta u$. Propagating the smoothness $\Delta u$ in the direction of isophotes $\nabla u^\top$, satisfies to the following equation [1]:

$$\nabla u^{*\perp} . \nabla \Delta u^* = 0 \quad \text{in} \quad \Omega \qquad (1)$$

$$u^* = u \quad \partial\Omega$$

Where $u^*$ : denotes image solution.

Equation (1) represents the basic model of inpainting and can be solved iteratively by introducing a diffusion time t :

$$\partial_t u = \frac{\partial u}{\partial t} = \nabla u^\perp . \nabla \Delta u \qquad (2)$$

$$u^{n+1} = u^n + \Delta t \ \nabla u^\perp . \nabla \Delta u \qquad (3)$$

Howerver, some problems exist such as the discontinuities preservation during diffusion and occlusion. The use of non linear partial differential equations (PDEs) approaches can provide reliable solution, which perform an anisotropic diffusion in a much better way.

## 3. The non linear diffusion PDEs formalism

In the past few years, the use of non linear PDEs methods involving anisotropic diffusion, has significantly grown and becomes an important tool in contemporary image processing. The key idea behind the anisotropic diffusion is to incorporate an adaptative smoothness constraint in the denoising or inpainting process. The diffusion process is governed by the following PDE [14]:

$$\partial_t u = div(g(|\nabla u|)\nabla u) \quad on \ \Omega \ x \ (0, \infty) \qquad (4)$$

$$u(x, y, 0) = u_0 \quad on \ \partial\Omega \ x \ (0, \infty)$$

Where $|\nabla u|$: denotes the gradient modulus and g(.) is a non-increasing function, known as the diffusivity function. This function allows isotropic diffusion in flat regions and no diffusions near edges. By developing the divergence term of (3), we obtain:

$$\partial_t u = g''(|\nabla u|)u_{\eta\eta} + g'(|\nabla u|)/|\nabla u|u_{\xi\xi} \qquad (5)$$

$$= c_\eta u_{\eta\eta} + c_\xi u_{\xi\xi}$$

Where: $u_{\eta\eta} = \eta^\perp H \eta$ and $u_{\xi\xi} = \xi^\perp H \xi$ are respectively the second spatial derivatives of $u$ in the directions of the gradient $\eta = \nabla u/|\nabla u|$, and its orthogonal $\xi = \eta^\perp$ ; H denotes the Hessian of u. According to these definitions, on the image discontinuities, we have the diffusion along $\eta$ (normal to edges) weighted with $c_\eta = g''(|\nabla u|)$ and a diffusion along $\xi$ (tangential to edges) weighted with $c_\xi = g'(|\nabla u|)/|\nabla u|$. Let now considering a contour **C**, as shown in figure 3, separating two homogeneous regions of the image, the isophote lines correspond to u(x,y) = c. The vector $\eta$ is normal to the contour **C**, the set $(\xi, \eta)$ is then a moving orthonormal basis whose configuration depends on the current coordinate point (x, y). Thus, for $c_\eta$=0, equation (5) describes a tangential diffusion weighted by $c_\xi$ along a contour, for a time t. Intuitively, the direction $\xi$ could interpolate existing structures within the missing areas.

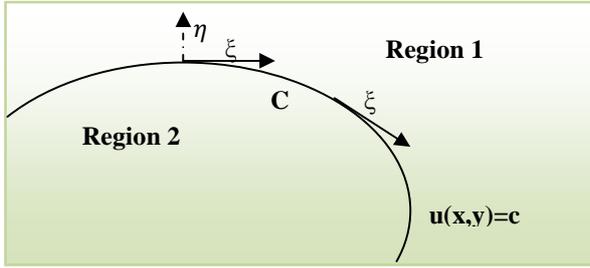

**Fig. 3** Diffusion along isophotes lines in the $\xi$ direction

An extension of the nonlinear PDEs to vector-valued image (e.g. color image) has been proposed [16]. It evolves $D \in R^n \to R^n$ under the diffusion equations:

$$\partial_t u_i = div(\ g(\sum_{k=1}^n |\nabla u_k|^2) \nabla u_i)\ \ i = 1..n \ \ on\ \Omega\ x\ (0,\infty) \quad (6)$$

Where $u_i$ : denotes the i$^{th}$ component channels of $u$.

Note that $\sum_{k=1}^n |\nabla u_k|^2$, represents the luminance function, which coupled all vector channels taking the strong correlations among channels. However, this luminance function is not being able to detect iso-luminance contours.

## 4. Diffusion Tensor approach

The non linear diffusion PDE approach, seen previously, does not give reliable information in the presence of complex structures images (e.g. fingerprint). It would be desirable to rotate the flow towards the orientation of interesting features. This can be easily achieved by using the structure tensor, also referred to the second moment matrix. For a multivalued image, the structure tensor has the following form [17]:

$$J_\rho = K_\rho * \left(\sum_{i=1}^n \nabla u_{i\sigma} \nabla u_{i\sigma}^\top \right) = \begin{bmatrix} j_{11} & j_{12} \\ j_{21} & j_{22} \end{bmatrix} \quad (7)$$

With :
$$\begin{cases} j_{11} = K_\rho * \sum_{i=1}^n u_{ix\sigma}^2 \\ j_{12} = j_{21} = K_\rho * \sum_{i=1}^n u_{ix\sigma} u_{iy\sigma} \\ j_{22} = K_\rho * \sum_{i=1}^n u_{iy\sigma}^2 \end{cases}$$

Where $\nabla u_{i\sigma} = K_\rho * \nabla u_i = K_\rho * (u_{ix}, u_{iy}) = (u_{ix\sigma}, u_{iy\sigma})$ : the smoothed version of the gradient obtained by convolving ; $K_\sigma$ and $K_\rho$ : denotes the Gaussian kernels with standard deviation $\sigma$ and $\rho$ respectively.

These new gradient features allow a more precise description of the local gradient characteristics.

To go further into the formalism, the structure tensor $J_\rho$ can be written over its eigenvalues ($\lambda_+$, $\lambda_-$) and eigenvectors ($\theta_+, \theta_-$), that is :

$$J_\rho = (\theta_+ \ \theta_-) \begin{bmatrix} \lambda_+ & 0 \\ 0 & \lambda_- \end{bmatrix} \begin{bmatrix} \theta_+ \\ \theta_- \end{bmatrix} = \lambda_+ \theta_+ \theta_+^T + \lambda_- \theta_- \theta_-^T \quad (8)$$

The eigenvectors of $J_\rho$ give the preferred local orientations, and the corresponding eigenvalues denote the local contrast along these directions. The eigenvalues of $J_\rho$ are given by :

$$\lambda_+ = \frac{1}{2}(j_{11} + j_{22} + \sqrt{(j_{11} - j_{22})^2 + 4j_{12}^2}) \quad 9)$$
$$\lambda_- = \frac{1}{2}(j_{11} + j_{22} - \sqrt{(j_{11} - j_{22})^2 + 4j_{12}^2}) \quad (10)$$

And the eigenvectors ($\theta_+, \theta_-$) satisfy :

$$\theta_- = \begin{pmatrix} -\frac{(j_{22} - j_{11} + \sqrt{(j_{11} - j_{22})^2 + 4j_{12}^2})}{\sqrt{(j_{22} - j_{11} + \sqrt{(j_{11} - j_{22})^2 + 4j_{12}^2})^2 + 4j_{12}^2}} \\ \frac{2j_{12}}{\sqrt{(j_{22} - j_{11} + \sqrt{(j_{11} - j_{22})^2 + 4j_{12}^2})^2 + 4j_{12}^2}} \end{pmatrix} \quad (11)$$

and $\theta_+ \perp \theta_-$.

The eigenvector $\theta_+$ which is associated to the larger eigenvalue $\lambda_+$ defines the direction of largest spatial change (i.e. the "gradient" direction). Correspondingly, the eigenvector $\theta_-$ is associated to the isophotes direction. We note that for the case of the scalar image (e.g. n=1), $\lambda_+ = |\nabla u|^2$, $\lambda_- = 0$, $\theta_+ = \eta = \nabla u/|\nabla u|$, $\theta_- = \xi = \nabla u^\perp/|\nabla u|$.

The eigenvalues ($\lambda_+, \lambda_-$) are indeed well adapted to discriminate different geometric cases. Weickert [17] proposed a non linear diffusion tensor by replacing the diffusivity function g(.) in (4) with a structure tensor, to create a truly anisotropic scheme, that is:

$$\partial_t u_i = div(D(J_\rho)\nabla u_i) \quad i = 1..n, on\ \Omega\ x\ (0,\infty) \quad (12)$$

Where $D(.)$ is the diffusion tensor which is positive definite symmetric 2x2 matrix.

This tensor possesses the same eigenvectors $\theta_-$, $\theta_+$ as the structure tensor $J_\rho$ and uses two weighted functions $\lambda_1$ and $\lambda_2$ to control the diffusion speeds in these two directions, that is:

$$D(J_\rho) = (\theta_+ \ \theta_-)\begin{bmatrix}\lambda_1 & 0 \\ 0 & \lambda_2\end{bmatrix}\begin{bmatrix}\theta_+ \\ \theta_-\end{bmatrix} = \lambda_1\theta_+\theta_+^T + \lambda_2\theta_-\theta_-^T \quad (13)$$

The diffusion tensor D takes the form of an ellipsoid as is represented in figure 4. We note that for the scalar image from (4), the diffusion tensor is reduced to $D = g(|\nabla u|)I_d$; where $I_d$ : Identity matrix.

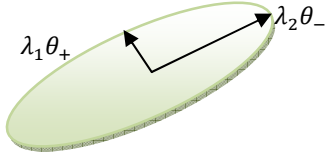

**Fig. 4** Diffusion tensor 2D representation

However, the choice of the two weighted functions λ1 and λ2 is very important to ensure adequate diffusion. For image denoising such as coherence enhancing diffusion (CED) [18], the eigenvalues are assembled via:

$$\lambda_1 = c_1 \quad (14)$$
$$\lambda_2 = \begin{cases} c_1 & \text{if } \lambda_+ = \lambda_- \\ else \ c_1 + (1-c_1)\exp\left(-\frac{c_2}{(\lambda_+ - \lambda_-)^2}\right) \end{cases}$$

Where $c_1 \in [0\ 1]$ and $c_2 > 0$.
- In flat regions, we should have $\lambda_+ = \lambda_- = 0$, and then $\lambda_1 = \lambda_2 = c_1$; $D = c_1 I_d$ where $I_d$ is the identity matrix. The tensor D is defined to be isotropic in these regions and takes the form of a circle of radius $c_1$.
- Along image contours, we have: $\lambda_+ \gg \lambda_- \gg 0$, and then $\lambda_2 > \lambda_1 > 0$. The diffusion tensor D is then anisotropic, mainly directed by the smoothed direction $\theta_-$ of the image isophotes.

From these results, it appears that the inpainting process requires diffusion along vector edges $\theta_-$.

## 5. Proposed Method

The proposed method stems from our earlier works [23] [24]. It is based on the diffusion tensor of equation (12). The idea is to avoid isotropic smoothing when diffusing image structure. As we mentioned previously, the diffusion should be done in $\theta_-$ direction of isophotes leading to coherent colors completions in the inpainted holes.

If we denote M : $R^2 \rightarrow \{0,1\}$ the mask function, we have :

$$\partial_t u_i = \text{div}(D\nabla u_i) = trace(DH_i) \text{ if } M=1 \quad (15)$$
$$\partial_t u_i = 0 \text{ if } M = 0$$

$H_i$: denotes the Hessian matrix
With the following proposed diffusion tensor:

$$D = f(\lambda_+, \lambda_-) \ \theta_-\theta_-^T \quad (16)$$

The function $f(.)$ ensures the regularization by weighting the diffusion process. We propose using the following function:

$$f(\lambda_+, \lambda_-) = c/\left(1 + \left(\frac{\sqrt{\lambda_+ + \lambda_-}}{k}\right)\right) \quad (17)$$

Where: c and k are two parameters which control the gain and the threshold tensor diffusion respectively. The choice of theses parameters depends on the size of the region to be filled. By developing equation (15) and introducing the directional derivative $u_{\theta_-\theta_-} = \theta_-^T H\theta_-$, we obtain:

$$\partial_t u_i = f(\lambda_+, \lambda_-) u_{i\theta_-\theta_-}, \ i = 1..n, \text{ if } M=1 \quad (18)$$

Furthermore, equation (18) can be solved numerically using finite differences. The time derivative $\partial_t u_i$ at $(i,j,t_n)$ is approximated by the forward difference $\partial_t u_i = (u_i^{n+1} - u_i^n)/\Delta t$, which leads to the iterative scheme:

$$u_i^{n+1} = u_i^n + \Delta t( f(\lambda_+, \lambda_-) u_{i\theta_-\theta_-}) \quad (19)$$

The main steps of the proposed algorithm are as follow:

- **Step 1:** *Parameters initialization: N, c, σ, ρ, k*
- **Step 2:** *Detect and Extract mask M from damaged region*
- **Step 3:** *while $s \leq N$ (iterations number)*
  *- Make discretization of the tensor components : $j_{11}, j_{12}, j_{21}, j_{22}$ of $J_\rho$ from eq.(7), by finite differences*
  *- Evaluate $\theta_-$ by using eq.(11)*
  *- Reconstruct $u_{i\theta_-\theta_-} = \theta_{i-}^T H\theta_{i-}$*
  *- Apply the iterative PDE scheme of (19) to the three color components (i=1..3)*

## 6. Experimental results

In the first experiment, we test and compare the performance of the proposed algorithm on a synthetically image showed in figure 5a which is characterized by a spirals orientations. The model's parameters are: $\Delta t=0.24$, $c = 0.75$, $k=0.050$, $\sigma=1.2$, $\rho = 4.5$, $N=2500$. The restored image in figure 5-f shows a good quality image, that is missing regions are perfectly filled with coherent colors.

The contours are well reconstructed and recovered with a good track of the local geometry image and isophotes. Compared to some existence methods such as: the Fast inpainting approach [13], the Total Variation [12] and the Harmonic inpainting [5], we note that they are not able correctly to interpolate missing edges with appearance of shading effect in hole region. To quantitatively evaluate the quality image, we use two measures: The PSNR and the mean structural similarity (MSSIM) index [19]. The PSNR is defined by:

$$PSNR = 10 log_{10} \frac{N_{max}}{MSE} \quad (20)$$

where $N_{max}$ : the maximum fluctuation in the input image, $N_{max} =(2^n-1)$, $N_{max} =255$, when the components of a pixel are encoded on 8 bits; MSE : denotes the mean square error, given by :

$$MSE = \frac{1}{MN}\sum_{i=1}^{N}\sum_{j=1}^{M}\left|f(i,j) - \hat{f}(i,j)\right|^2 \quad (21)$$

Where f(i,j) : the original image, $\hat{f}(i,j)$ : the restored image.

The MSSIM approximates the perceived visual quality of the image better than PSNR. The MSSIM takes values in the interval [0,1] and increases as the quality increases. Table 1 confirms the effectiveness of our model with the highest score of both PSNR and MSSIM. Figure 6 shows the results of a damaged photographic image. The model's parameters are fixed to: $\Delta t =0.24$, $c = 0.75$, $k=0.180$, $\sigma=1.2$, $\rho = 4.5$, $N=2500$. The restored image in figure 6.f, is well repaired with a good score of PSNR and MSSIM as shown in table 2. The target regions are reasonably filled without color artifacts appearance.The algorithm maintains the continuity of isophotes directions while reducing significantly the over-smoothing effect, which is clearly perceived for the others methods.

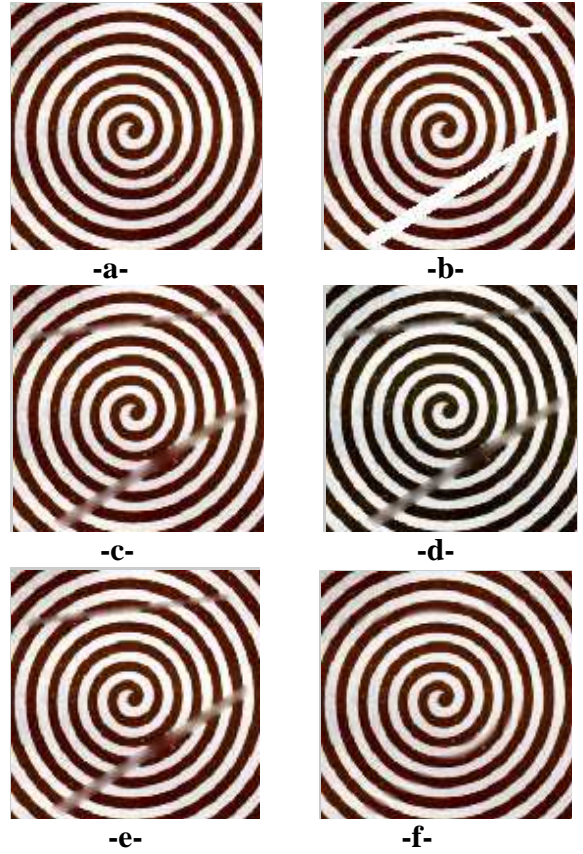

**Fig. 5** *Test on synthetically image,-a- Original image, -b- Damaged image,-c-TV inpainting method, -d- Fast Inpainting method -e- Harmonic inpainting method,-f- Proposed method*

Table 1

| Methods | PSNR(dB) | MSSIM |
|---|---|---|
| Fast inpainting | 19.49 | 0.943 |
| TV inpainting | 21.81 | 0.951 |
| Harmonic inpainting | 20.82 | 0.948 |
| Proposed method | 28.20 | 0.982 |

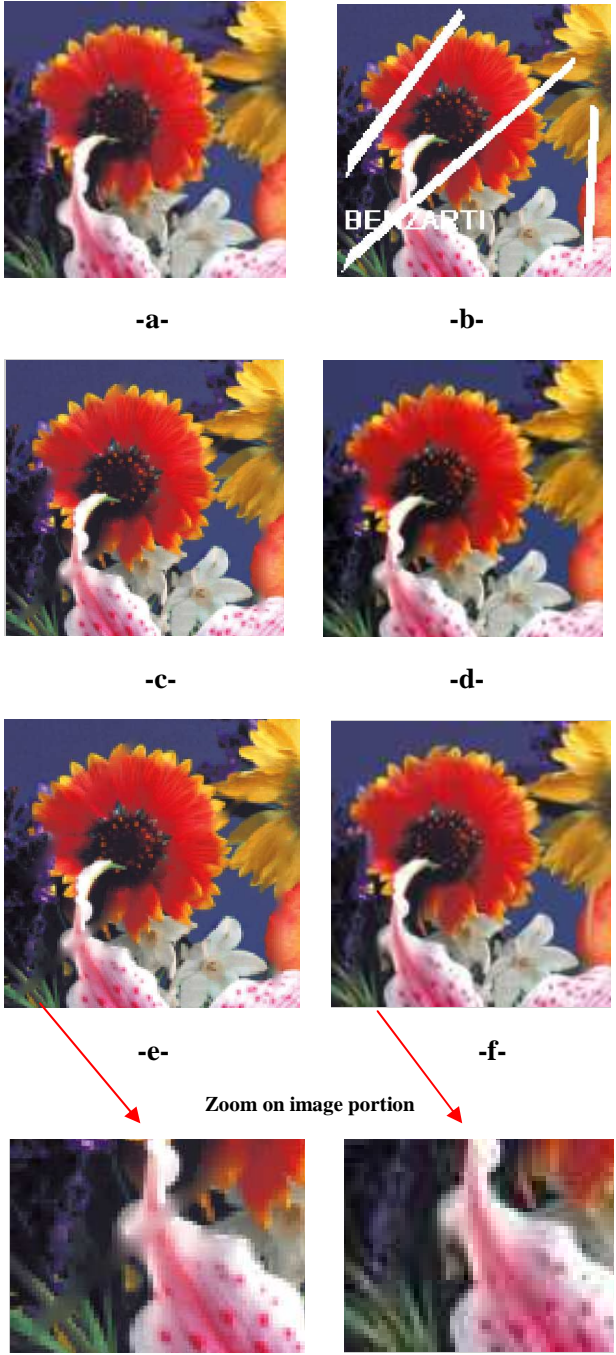

**Fig. 6** *Test on real photographic image,-a- Original image, -b- Damaged image,-c-TV inpainting method, -d- Fast Inpainting method -e- Harmonic inpainting method,-f- Proposed method*

Table 2

| Methods | PSNR(dB) | MSSIM |
|---|---|---|
| Fast inpainting | 31.92 | 0.965 |
| TV inpainting | 32.08 | 0.968 |
| Harmonic inpainting | 31.78 | 0.963 |
| Proposed method | 33.78 | 0.978 |

It should be noted that a small problem of occlusion appears when two colors image are superimposed. Furthermore, the efficiency of any methods in the structure inpainting category depends on the area of the inpainting region. If this missing region is small and the surrounding area is without much texture, the result will be acceptable subjectively.

## 7. Conclusion

In this paper, we have proposed an efficient method for image inpainting based on the non linear diffusion tensor. This later is more appropriate than the use of a simple diffusion gradient since it gives reliable information in the presence of complex structures images. The algorithm performs inpainting by joining with geodesic curves the points of the isophotes arriving at the boundary of the region to be inpainted. Experimental results on test and real digital pictures are very promising in terms of repairing damaged zones and discontinuities preservation. Future work will include texture inpainting for large damaged zones and developing method for occlusion problem.

**Faouzi  Benzarti**  is an associate professor at the High School of Technology and Science of Tunisia (ESSTT). He received his Engineer's degree in Electrical Engineering from the Engineering School of Monastir (TUNISIA) in 1987, and his master's degree in Biomedical Engineering from the Polytechnic School of Montreal CANADA in 1991. He obtained his Ph.D degree from the Engineering School of Tunis (ENIT) in 2006.  He is a member of research group in the Signal, Image and Pattern Recognition Laboratory TSRIF. His current researches include: Image Deconvolution, Image Inpainting, Anisotropic diffusion, Image retrieval, 3D Biometry.

**Hamid Amiri**  is  a Professor at the National Engineering School of Tunis (ENIT) Tunisia. He received the Diploma of Electro-technics, Information Technique in 1978 and the PHD degree in1983 at the TU Braunschweig, Germany. He obtained the Doctorates Sciences in 1993. From 2001 to 2009 he was at the Riyadh College of Telecom and Information. Currently. He is now thea head member of research group in the Signal, Image and Pattern Recognition Laboratory. His research is focused on Image Processing, Speech Processing, Document Processing and Natural language processing.